\documentclass[a4paper,10pt,onecolumn,draftclsnofoot,]{IEEEtran}

\IEEEoverridecommandlockouts                              

\usepackage{graphicx}

\usepackage{algorithm,algpseudocode}
\algnewcommand{\To}{\textbf{To }}
\algnewcommand\Input{\item[\textbf{Input:}]}%
\algnewcommand\Output{\item[\textbf{Output:}]}%
\algnewcommand\Parameter{\item[\textbf{Parameters:}]}%

\usepackage{subfig}
\usepackage{amsfonts}
\usepackage{gensymb}
\usepackage{amsmath,amssymb,amsfonts}
\usepackage{url}
\usepackage{etoolbox}
\usepackage{graphicx}

\usepackage{geometry}
\usepackage{fancyhdr}

\fancypagestyle{mystyle}{%
    \fancyhead[CO, CE]{This paper has been accepted for publication at IROS 2023. \\ Please cite the paper as: M. Stamatopoulos, A. Banerjee, G. Nikolakopoulos, 2023 IEEE/RSJ International Conference on Intelligent Robots and Systems (IROS 2023)}
}%

\title{\LARGE \bf
Flexible Multi-DoF Aerial 3D Printing Supported with Automated Optimal Chunking}

\author{Marios-Nektarios Stamatopoulos$^{1}$, Avijit Banerjee$^{1}$ and George Nikolakopoulos$^{1}$
\thanks{$^{1}$The Authors are with the Robotics and Artificial Intelligence Group, Department of Computer, Electrical and Space Engineering, Lule\r{a} University of Technology, 971 87 Lule\r{a}, Sweden}%
\thanks{Cooresrponing author's e-mail: \tt\small{marsta@ltu.se}}
}

\begin{document}
\maketitle
\thispagestyle{mystyle}

\begin{abstract}
The future of 3D printing utilizing unmanned aerial vehicles (UAVs) presents a promising capability to revolutionize manufacturing and to enable the creation of large-scale structures in remote and hard-to-reach areas e.g. in other planetary systems. Nevertheless, the limited payload capacity of UAVs and the complexity in the 3D printing of large objects pose significant challenges. In this article we propose a novel chunk-based framework for distributed 3D printing using UAVs that sets the basis for a fully collaborative aerial 3D printing of challenging structures. The presented framework, through a novel proposed optimisation process, is able to divide the 3D model to be printed into small, manageable chunks and to assign them to a UAV for partial printing of the assigned chunk, in a fully autonomous approach. Thus, we establish the algorithms for chunk division, allocation, and printing, and we also introduce a novel algorithm that efficiently partitions the mesh into planar chunks, while accounting for the inter-connectivity constraints of the chunks. The efficiency of the proposed framework is demonstrated through multiple physics based simulations in Gazebo, where a CAD construction mesh is printed via multiple UAVs carrying materials whose volume is proportionate to a fraction of the total mesh volume.
\end{abstract}
\section{Introduction}
The recent advances in three-dimensional ($3$D) printing technologies are expected to revolutionize the capabilities consideration of the construction and manufacturing sectors~\cite{karasik2019object,balletti20173d}. Until now, the maturity of additive manufacturing has established its ubiquitous outreach in the fields of healthcare~\cite{zuniga2015cyborg}, agriculture~\cite{pearce2015applications}, construction~\cite{craveiroa2019additive}, automotive and aerospace industries~\cite{joshi20153d} to name a few. Currently, there is a trend where researchers are looking forward to the feasibility of the $3$D printing in enabling the construction of real-scale infrastructures for building e.g. emergency shelters, providing relief for post-disaster accommodations in remote, hostile and hard-to-access environments exposed to extreme climates, unstructured terrain, and distant military locations~\cite{al2018large,bazli20233d}. In principle, a 3D printer builds any physical object directly from a computer-aided design (CAD) model by successively depositing materials in a stereolithography fashion~\cite{zhao2022general}. However, the size of a built object is restricted by the dimensions of the associated machinery, limiting the shape of the printable structure, confined within the constraint of the manufacturing envelope. Hence, it is challenging to realize the envisioned objective of scalable construction with state-of-the-art technological support, which demands extensive infrastructure that is unrealistic for highly remote and difficult-to-access areas  

\begin{figure}[htbp]
    \centering
    \includegraphics[width=0.6\textwidth]{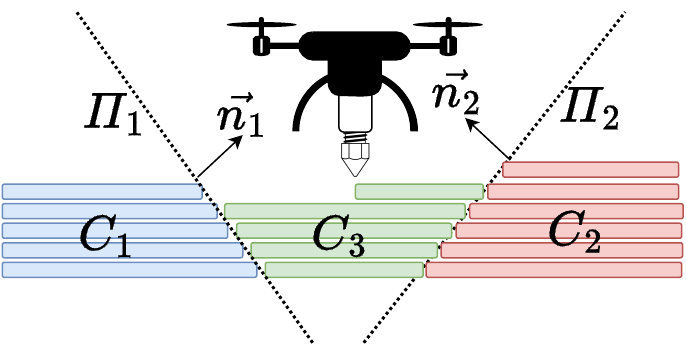}
    \caption{Concept of Aerial 3D Printing UAV with mesh decomposed into chunks through planar cuts}
    \label{fig:my_label}
\end{figure}

In this context, the concept of aerial $3$D printing is recently proposed as the viable alternative~\cite{zhang2022aerial}. The conceptual framework, inspired by natural builders, such as wasps and bees, considers a group of Unmanned aerial vehicles (UAVs) collaboratively participating to autonomously construct a 3D structure in a controlled manner. Thanks to the significant advancements in autonomous UAV technologies that have the potential to enable precise motion planning, navigation and task execution for UAVs, a systematic feasibility study for drone-based masonry construction of real scale structure was presented in ~\cite{goessens2018feasibility,hunt20143d}. Thus, the utilization of aerial robots that deposit materials for $3$D printing is innovative and still a relatively unexplored area that is evolving as a promising direction for autonomous scalable constructions. Until now, very few research approaches have been reported, which are primarily in their primitive conceptual stage. In this article, a more generic and flexible approach for aerial $3$D printing is presented as shown in concept Fig. \ref{fig:my_label}, targeting seamless integration for the simultaneous operation of multiple aerial robots in a collaborative framework.

\subsection{Related work}
The aerial $3$D printing, presented so far, considers a primitive design strategy, which typically replicates the classical additive manufacturing architecture that is built on a $2.5$ dimensional formalism. Hence, the motion of aerial robots, and the associated path of extruders, are primarily confined within horizontal ($X-Y$) planes with limited variability along the vertical direction ($Z$-axis)~\cite{zhang2022aerial,hunt20143d}. An extensive review reported in~\cite{jiang2021review} reveals that such restricted manoeuvrability leads to poor bonding strength between layers, staircase effects on surface quality, and requirement for an external support. In view of that, flexible multiple DoF motion planning is recommended as an advanced printing strategy. 

In this context, considering the agile $6$-DOF manoeuvrability of aerial robots, more flexible motion planning and seamless coordination can be exploited for fast construction of complex shapes with minimum aid of external support structure, which is a critical factor for $3$D printing in general~\cite{ruan2007automated}. Typically, the $3$D printing requires
auxiliary supporting structures to prevent the growing objects from deforming and even collapsing under the effect of gravity. In order to realize a support-free construction, with the aid of multiaxis layered manufacturing, various optimal slicing mechanisms are considered~\cite{ChopperΜΙΤ,gao2019near} that decompose the geometric shape of the object to be printed into multiple sub-parts called chunks. In the context of aerial $3$D printing, the chunks need to be designed to incorporate printability constraints adhering to the shape of the extruder, while enabling provision for smooth connectivity between sub-parts. In view of collaborative manufacturing, a scheduling mechanism \cite{wenchao_zhou_ground_robots} determines the sequence of chunks that needs to be printed in a distributed fashion to construct the object in a bottom-up approach. However, to the best of our knowledge, such an advanced generic framework involving multiaxis-layered manufacturing in the context of aerial $3$D printing has not been reported in the literature.

\subsection{Contributions}
In this article, a novel generic collaborative aerial $3D$ printing framework is presented that enables seamless integration of distributed aerial $3D$ printing with multiple UAVs in a multiaxis layered manufacturing framework. Thus the main contributions of this article stem from the introduction of a generic chunking mechanism in the context of collaborative aerial $3$D printing that optimally decomposes an arbitrary geometric shape (to be printed) into multiple chunks. The second contribution stems from the fact that the chunks are generated based on planner cuts, incorporating printability constraints and adhering to the shape of the extruder, while enabling provision for smooth connectivity between adjacent sub-parts. The third contribution is based on the establishment of an innovative task scheduling mechanism, which produces a feasible sequence and priorities for parallel execution of distributed $3$D printing in a coordinated manner and specifically suited for multiple UAVs carrying aerial printing capabilities. Finally, the overall evaluation of the proposed aerial $3$D printing mechanism is performed in a GAZEBO based simulation framework incorporating multiple UAVs in printing a complex 3D structure of large scale.
\subsection {Organization of the Article}
The rest of the article is structured as follows. In Section \ref{Sec:Problem formulation}, an overview of the problem formulation is presented, while in Section \ref{Sec:Chunk Generation}, the novel mechanism for the construction of the feasible chunks is proposed. The execution of aerial printing with UAVs and associated trajectory generation and a path-following controller is presented in Section \ref{Sec:UAV Execution}. The overall efficiency of the proposed framework is validated with realistic simulation results in Section \ref{Sec:Simulation Results}, while the article is concluded in Section \ref{sec:conclusions}. 
%
\section{Problem Formulation} \label{Sec:Problem formulation}
\begin{figure*}[htbp]
    \begin{center}
        \includegraphics[width=\textwidth]{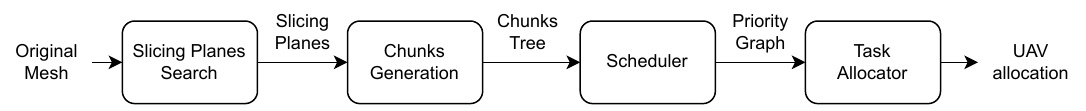}
        \caption{System Block Diagram} 
        \label{fig:block_diagram}
    \end{center}
\end{figure*}

Towards making aerial 3D Printing viable, a mesh decomposition procedure is presented, which distributes the original geometric shape to be printed into optimally selected multiple smaller sub-meshes called chunks. This is achieved by dividing the initial mesh into multiple planes of different orientations. Hence, the problem of mesh decomposition is transformed into searching for those planes that will distribute the geometric shape into a set of the best possible combinations of chunks. The planes are uniformly sampled, while imposing a beam search algorithm that keeps the most promising cuts in every iteration. Each set of cutting planes will result in chunks of the original mesh. A heuristic is assigned, which evaluates the uniformity of distributed volume over generated chunks. While this search is being executed, a Binary Space Partitioning Tree (BSP) \cite{BSPTree} is constructed, containing all the cuts and the resulting structure of the produced chunks. The BSP tree helps in keeping a trace of a sequence, which can be used to trace back the geometric shape, while reconstructing the original structure. Using the BSP tree, a scheduler is constructed that transforms the tree into associated dependencies between chunks which describes a priority sequence. In order to regenerate the expected original mesh, the priority sequence needs to be followed while printing. An overview of the entire procedure is presented schematically in Fig. \ref{fig:block_diagram}, while a systematic, detailed description of each block is presented next.
%
%
%
\section{Chunks Generation} \label{Sec:Chunk Generation}
The construction mechanism responsible for the formation of the chunks generated based on planar cuts is presented in this Section. It is assumed that only cuts with planes of the form $\mathbf{\Pi} (\vec{n}, \vec{p})$ are allowed, where $\vec{n},\vec{p}$ are the normal vector and a point on it. Moreover, a BSP tree $\mathbf{T}$ \cite{BSPTree} is considered to represent the original mesh and its chunks throughout the entire execution of the framework. 

\begin{figure}[htbp] 
    \centering
    \includegraphics[width=0.6\columnwidth]{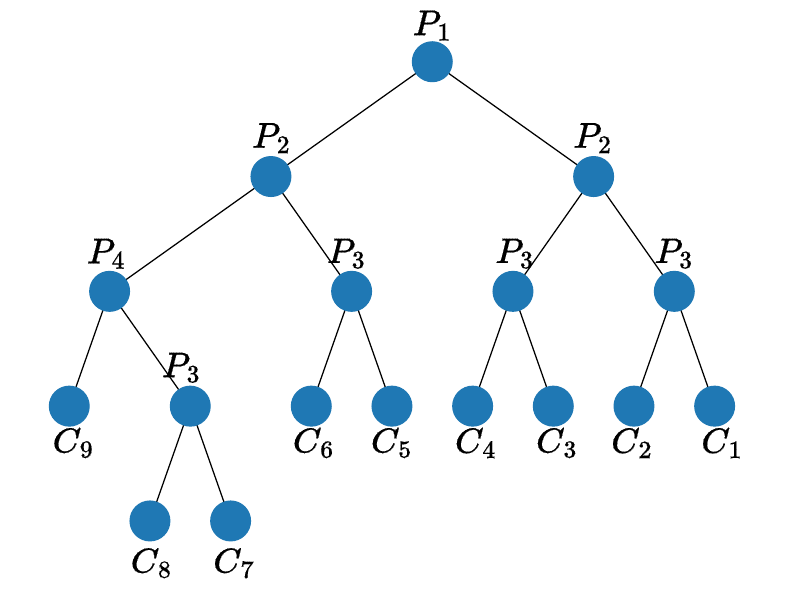}
    \caption{Tree Interpretation of all the cuts resulting into the chunks. Each node represents a new cut while all the chunks can be found on the leaf nodes}
    \label{fig:BSPtree}
\end{figure}

Each time a plane $\mathbf{\Pi}$ is given, and a cut is demanded, the leaf sub-part that intersects with this plane is found, and the cut is executed, which leads to two sub-parts. The sub-part that lies in the direction of the normal vector is considered positive and is placed as the right subordinate. Correspondingly the opposite sub-part is considered negative and placed on the BSP tree as the left subordinate. The node that was previously representing the leaf sub-part is now considered a new planar cut containing the plane $\mathbf{\Pi}$. The aforementioned procedure is executed recursively for every cut and leads to a tree structure representation as shown in Fig. \ref{fig:BSPtree}.
\subsection{Searching}
The primary functionality of the Chunker module is to find the planes that can be utilized as planar cuts in the 3D mesh to break it down into smaller pieces called "chunks". 
Similar to \cite{ChopperΜΙΤ}, the Chunker examines a list of the previously generated BSP trees in each iteration and tries to extend them by adding new cuts. It retains only the most promising options, as determined by the beam width (which is essentially a tuning parameter), and discards the rest. The selected options are then incorporated into the list of implemented cuts to create new sub-meshes. This approach allows the Chunker module to efficiently search for the most appropriate planar cuts and break the mesh into smaller and more manageable components. The search procedure for finding the final set of cuts can be shown in Algorithm \ref{alg:planesSearch}. 

            
    

\begin{algorithm}[H]
    \caption{Plane Cuts Search}\label{alg:planesSearch}
  \begin{algorithmic}[0]
      \State $srchTrees \gets  [mesh]$
      \While{not allTreesFeasible(srchTrees)}
          \State nonFeasibleTrees $\gets$ getNonFeasible(srchTrees)
          \State $newTrees \gets [~]$
          \For {\textbf{each} $tr$ \textbf{in} $nonFeasibleTrees$}
            \State $nonFeasibleTrees$.pop($tr$)
            \State $trees \gets$ evlauatePossiblePlanes($tr$)
            \State $trees \gets $ sorted($trees$, $tree$.cost)[:$W_{inner}$]
            \State $newTrees$.extend($trees$)
        \EndFor
        \State $newTrees\gets$sorted($newTrees$,tree.cost)[:$W_{outter}$]
        \State $srchTrees$.extend($newTrees$)
      \EndWhile
      \State \Return sorted($srchTrees$,$tree$.cost)[0]
  \end{algorithmic}

\end{algorithm}
    
Throughout the execution of the search algorithm, a list of the best trees is kept track of. All of them that do not satisfy the termination condition is chosen to be further expanded by adding an extra cut. The process carried out in order to both calculate and evaluate the possible extension cuts for the tree can be seen in Algorithm \ref{alg:evalCuts}.
\begin{algorithm}[H]
    \caption{Evaluate Possible Extensions for Tree}\label{alg:evalCuts}
    \begin{algorithmic}[0]
        \Input{$givenTree$}
        \Parameter{$angleLimits$,$n$} 
        \State$sampledNormals \gets$sampleNormals($angleLimits$,$n$)
        \State $newTrees \gets [~]$
        \For {\textbf{each} $normal$ \textbf{in} $sampledNormals$}
            \State $planes \gets$ getCoorespPlanes($n$)
            \For {\textbf{each} $p$ \textbf{in} $planes$}
                \State  $tr \gets$ $givenTree$.sliceWithPlane($p$)
                \State  $tr$.evaluateHeuristic()
                \State $newTrees$.append($tr$)
            \EndFor
        \EndFor
      \State \Return $newTrees$
  \end{algorithmic}
\end{algorithm}



\begin{figure}[H] \label{fig:sampleNormals}
    \centering
    \includegraphics[width=0.530\columnwidth]{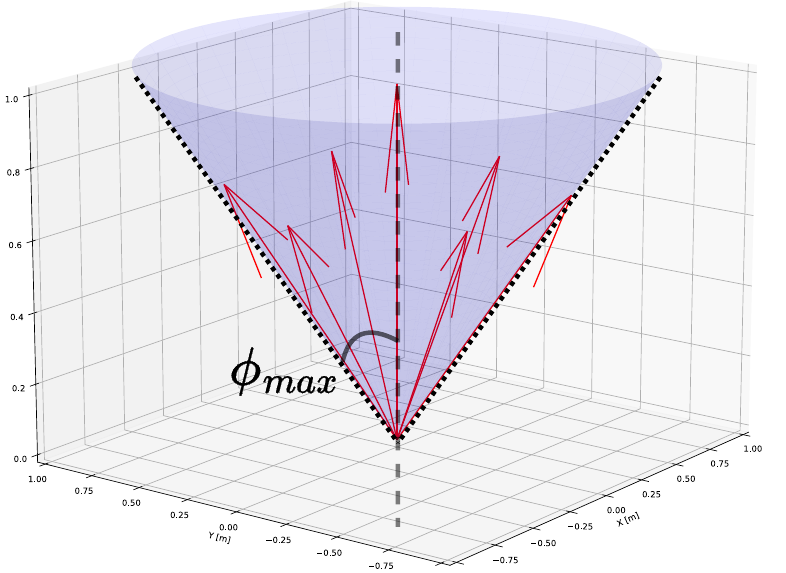}
    
    \caption{A schematic representation depicting the distribution of few Sampled Normals over an arbitrary geometric shape}
    \label{fig:unstable_traj_gen} 
\end{figure}

For every tree, a collection of different cutting planes, 
representing new possible cuts that will extend it, is 
calculated. The set $\mathbf{P}=\{ \mathbf{P_i}\}$ of possible 
cut planes is sampled. The problem constraints are taken into 
consideration during the sampling of the planes. The normal 
vectors set $\mathbf{N}=\{ \vec{n_i} \}$ (shown in Fig. \ref{fig:unstable_traj_gen}) of the planes lay 
along the spherical surface $\mathbf{S} \in \mathbf{R^3}= \{r, 
\theta,\phi \}$ where $r=1$, $\theta \in [0,2\pi)$ and 
$\phi \in [-\phi_{max}, \phi_{max}]$ represented in spherical 
coordinates.
The normals are sampled with a fixed angle step 
depending on the constant $M$ representing the number of 
planes that need to be sampled. 

\begin{figure}[h]
    \centering
    \includegraphics[width=0.50\columnwidth]{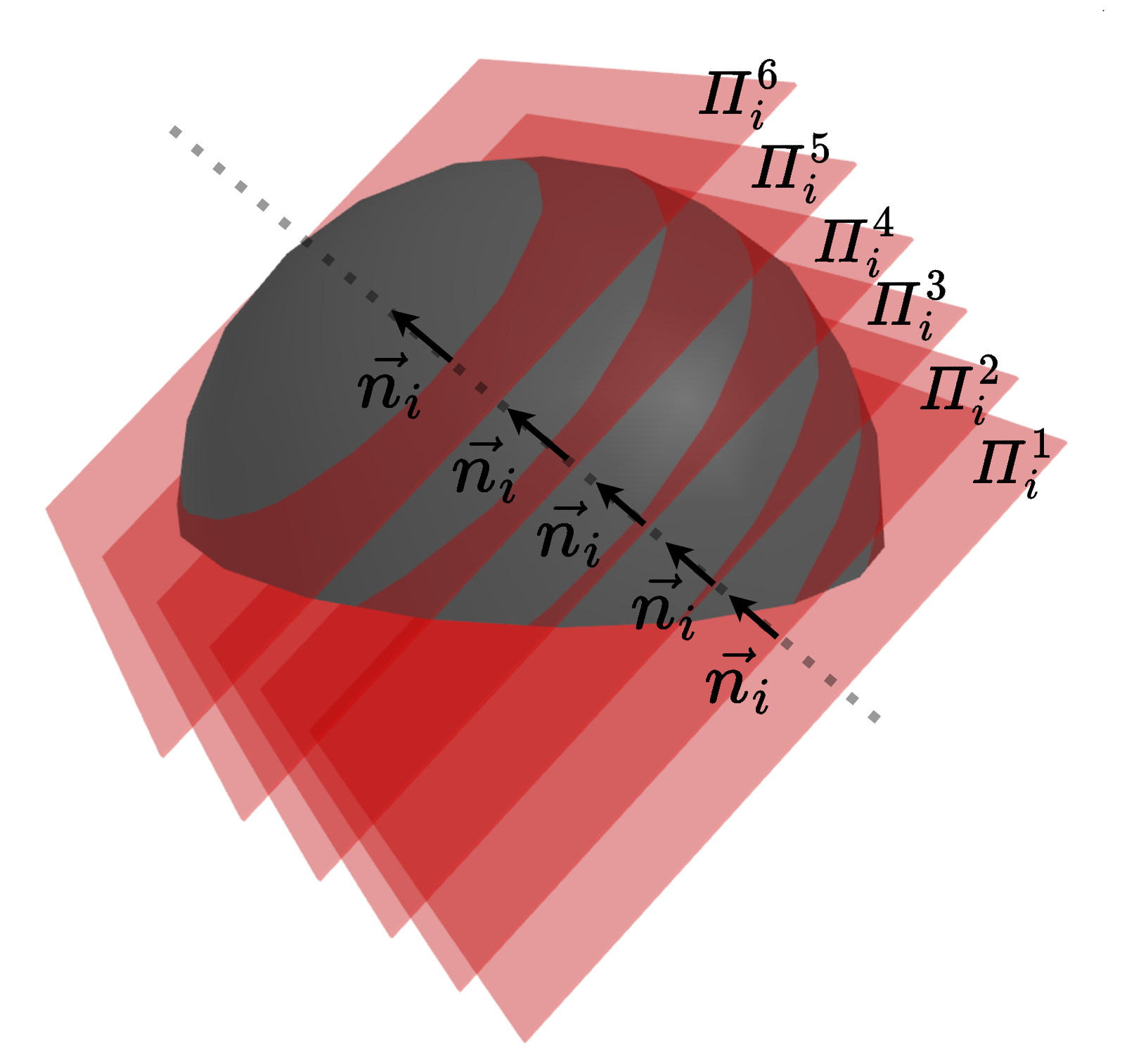}
    \caption{A conceptual demonstration depicting a family of planes obtained for a choice of sampled normal vector $\vec{n_i}$}
    \label{fig:planeFamily}
\end{figure}

For every sampled normal $\vec{n_i}$, a family of planes $\mathbf{F}=\{ \Pi_i^j\}$ is calculated. This is achieved by inserting an offset $d_i \in \mathbb{R}$ and selecting the position $\vec{p_j}= j * \vec{ni}$ of the plane to be along the direction of the normal vector $\vec{n_i}$, where $j$ is bounded by calculating all the projections of the mesh vertices $v_k$ along the normal vector $\vec{n_i}$. The set $\mathbf{M} = \{j \in \mathbb{R}: j=proj(v_k,\vec{n_i})\}$  contains all of them.
Finally, each plane of the family is defined as $\Pi_i^j= (\vec{p_j}, \vec{n_i})$, a representative scenario with a family of planes is presented in Fig. \ref{fig:planeFamily}.

%

For every plane sampled during the above process, a new cut is made, and the tree is extended. The re-evaluation of the tree is carried out based on the heuristics imposed and saved to each tree. From all the sampled expansions for each tree, only the $W_{inner}$ best ones are chosen to advance towards the next stage of the search. The $W_{inner}$ is usually set to a small value, which is a tuning parameter which balances the trade-off between intensive exploration and computational load.

Following the above-mentioned procedure, for all the trees that do not satisfy the terminal condition, are led to a bigger list containing all the newly expanded trees. The best $W_{outer}$ of those are kept based on their heuristic value and the rest are discarded so that the search can be focused on the most promising candidates, while still exploring a diverse range of possibilities and keeping the computational load low since a full search on all the expansion would lead to an exponential growth of candidates and the need for the complete computational geometry calculations carried out throughout the cutting and evaluating would be intense. 
\subsection{Heuristic}
The process of calculating the heuristic for a tree can be broken down into a sum of various sub-heuristics that aim to penalize different objectives. One of the objectives of the search process is to result in chunks with uniformly distributed volumes. Assuming that the overall volume of the original mesh is $\mathbf{V}$ and the $i$-th chunk's volume is notated as $V_i$ then $\mathbf{V}= \sum_{i=1}^{N} V_i$, where $N$ is the number of chunks. The coefficient of variation $c_v$ is chosen as a measure of the dispersion of the chunk volumes. Specifically, the standard deviation of the volumes $\sigma = {\sum_{i=1}^{N} (V_i - \mu)^2}/{N}$ along with the mean $ \mu=\sum_{i=1}^{N} V_i/{N}$ are calculated, and the final heuristic is presented as follows
\begin{align} \label{eq:dispercionCost}
    c_v = \frac{\sigma}{\mu} 
\end{align}
The following constraints are considered while constructing the chunks 
\subsubsection{Chunks Connectivity}
\begin{figure}[htbp]
    \centering
    \includegraphics[width=0.65\columnwidth]{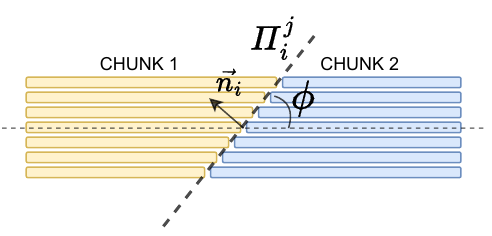}
    \caption{Connectivity constraint between chunks}
    \label{fig:conn_constraint}
\end{figure}

    The search process generates chunks that are going to be printed one on top of each other. In order to make this feasible, there must be sufficient coverage area between the chunks so that the adjacent chunks can stick to each other, as shown in Fig. \ref{fig:conn_constraint}. This constraint is integrated into the sampling step of the cutting planes. The angle $\phi_{max}$ (depicted in Fig.\ref{fig:unstable_traj_gen}) imposes the maximum angle of the plane normal with the $z$-axis, which determines the slope of the adjacent face of the two chunks. So, the absolute limit of this angle is chosen to be $\phi^{conn}_{max} = 45 \degree $, which is a valid assumption made in many applications of 3D Printing where it is used as a threshold for printing support structures or not. It must be noted that this value may differ on the material used for printing, and it is up to the user to set it based on the specifications and the objectives they want to achieve.
\subsubsection{Extruder Head Collision}
\begin{figure}[H] 
    \centering
    \includegraphics[width=0.65\columnwidth]{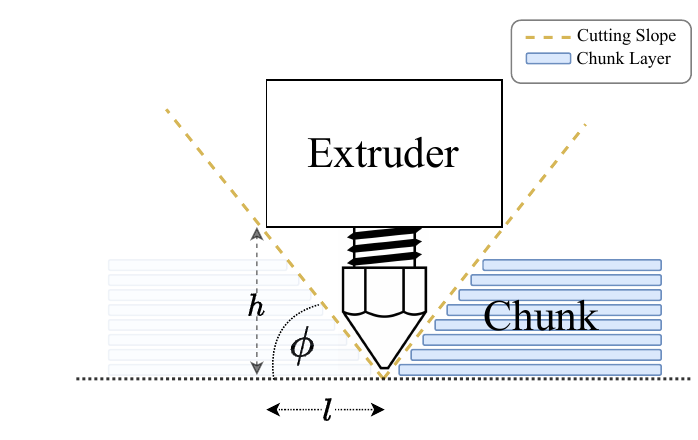} \caption{Extruder Head Collision Constraints}
    \label{fig:extruderHeadConstraint}
\end{figure}

Another constraint that is taken into account is the possible collision between the extruder head/nozzle and a previously printed chunk. As seen in Fig. \ref{fig:extruderHeadConstraint}, the maximum cutting slope depends on the geometry of the extruder nozzle and head. 

By assuming that the extruder's head is rectangular, the maximum value angle $\phi$ can be calculated knowing the length $l$ from the extruder nozzle to the extruder's head outer face and the height $h$ of the extruder nozzle. Then, the constraint can be simply defined as follows
\begin{equation}
    \phi^{extr}_{max} \leq arctan(\frac{h}{l})
\end{equation}

Finally, the biggest angle from the above is chosen as the overall constraint angle and will be considered as the maximum angle that a planar's cut normal can have.
\begin{equation}
    \phi_{max} = max(\phi^{conn}_{max}, \phi^{extr}_{max})
\end{equation}
\subsection{Feasibility Condition}
%
The initial configuration of the UAVs is known a priori and is denoted as $\mathbf{D} = [d_1, d_2, \dots, d_n]$, where $d_i$ is the material volume carried by the $i$-th UAV and $n$ the number of the available UAVs. The configuration is sorted in a descending order so that $d_i>d_{i+1}$, $\forall i \in [1,\dots,n]$. The same procedure is followed for the volumes of the chunks, which are defined by the set $\mathbf{C}=[c_1,c_2, \dots,c_k]$. 

\begin{figure}[H]
\begin{center}
    \subfloat[Original Mesh]{{\includegraphics[width=6.0cm]{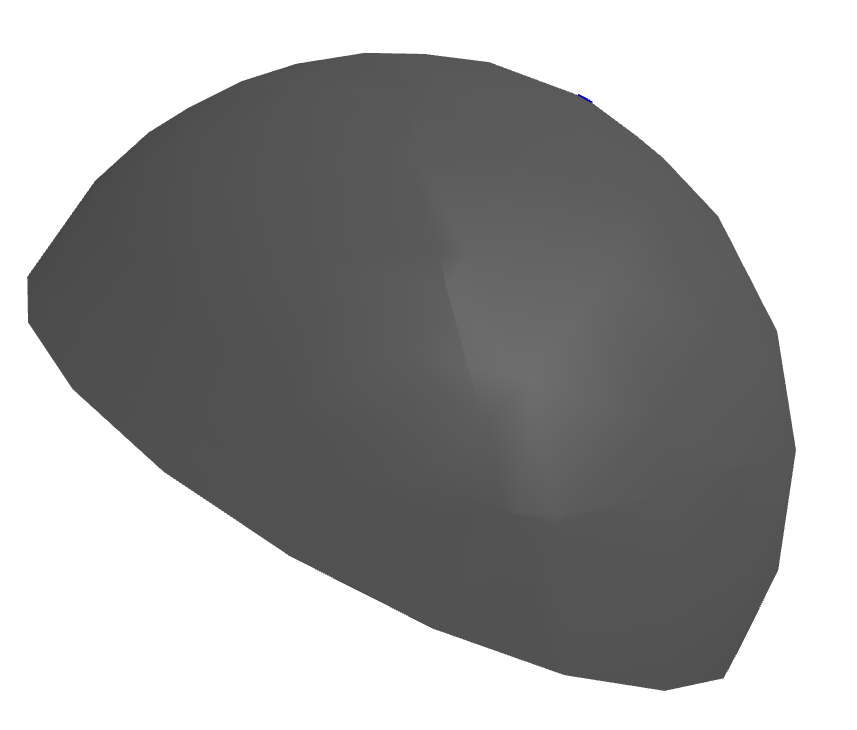} }}
    \qquad
    \subfloat[Chunks Generated]{{\includegraphics[width=6.0cm]{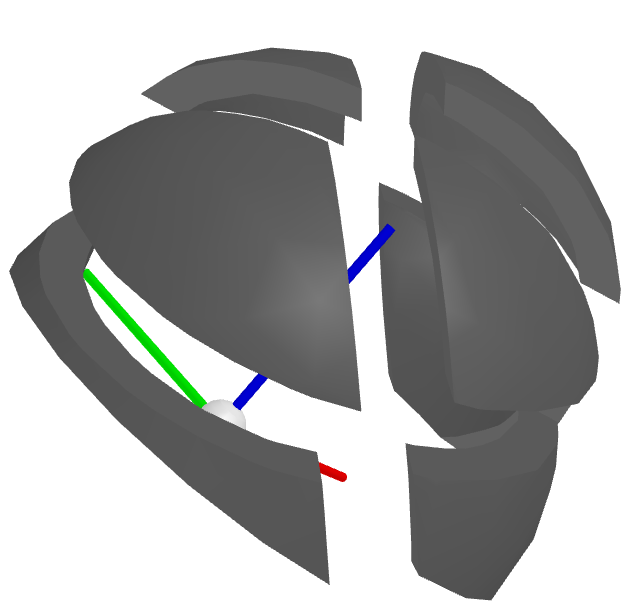} }}
    
    \caption{Chunks generated after planes search}
    \label{fig:search_results}
\end{center}
    
\end{figure}

Before any execution of the search algorithm, a primal feasibility condition must be satisfied that ensures that the available material being carried by the UAVs is greater than the volume of the whole original mesh. Let $V$ be the volume of the original mesh, then the following condition must be satisfied:
\begin{equation} \label{eq:feasibilityCond}
   V \leq \sum_{i=1}^{n} d_i
\end{equation}
A tree is feasible to be printed when there exists a non-empty set defined as:
\begin{eqnarray} \label{eq:term_condition}
    S = \{p_l= (d_j,c_i) :  d_j>c_i\} \neq \emptyset, \forall l=[1,\dots,k]
\end{eqnarray}
After the successful manufacturing of each chunk, $C, D$ and $S$ are considered to be updated recursively.  A mesh decomposition result for a hemispherical dome is shown in Fig. \ref{fig:search_results}.

\subsection{Chunks Priority}
The BSP tree mechanism, used for handling the planar cuts throughout the search phase, is having the advantage of inherently defining the priority that the chunks need to be stacked in order to be connected and fully materialize the original mesh given as input. Due to the constraints of the search, all the plane normals are facing the upper half space defined by the plane having as a result the positive mesh laying on top of the negative one, imposing that the negative needs to be printed first. This property can be recursively expanded and since all the negative chunks are placed as left children to the node representing the cut, then the chunk sequence that needs to be printed can be calculated by executing an in-order traversal on the leafs of the final BSP Tree $\mathbf{T}$ extracted by the search phase.



\section{UAV EXECUTION} \label{Sec:UAV Execution}
\begin{figure*}[h]
    \begin{center}
        \includegraphics[width=\textwidth]{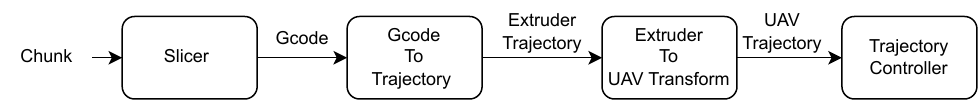}
        \caption{UAV Trajectory generation from given chunk } 
        \label{fig:UAVblock_diagram}
    \end{center}
\end{figure*}

After a chunk is assigned for printing, a series of steps need to be followed in order to actually print the Chunk, a block diagram of the associated sequence is presented in Fig. \ref{fig:UAVblock_diagram}.

\subsection{Chunk Slicer}
Initially, the mesh of the chunk that is a set of vertices and triangles is given to the slicer. This module's output is the path the extruder must follow in order to print the input mesh. This path is calculated using the commercial open-source software "Cura" \cite{cura}. The physical properties of the UAV can be taken into account by adjusting the input parameters to the slicer like layer height between layers, printing line width and infill percentage. By fitting these values to the corresponding use-case and UAV setup the path the extruder of the UAV needs to follow is extracted along with the details for the exact segments where material needs to be deposited and if yes, its extrusion rate as well. The format of the G-code is only segments connected with each other either in a linear fashion or by utilizing predefined curved paths. Though, this format can not be handle from the UAV controller and needs to be transformed.

\subsection{G-Code to UAV Trajectory}
\begin{figure}[H]
    \centering
    \includegraphics[width=0.4\columnwidth]{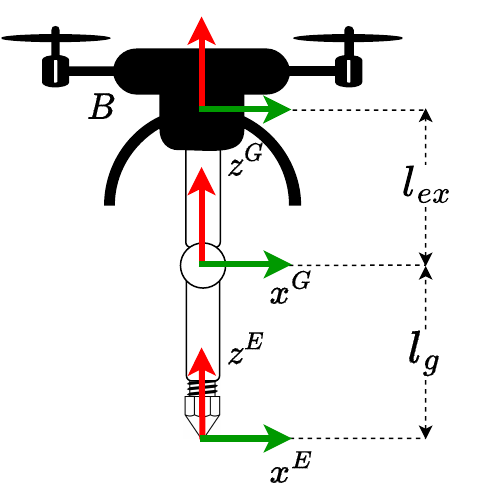}
    \caption{UAV Extruder Setup Side View}
    \label{fig:uav_extruder}
\end{figure}

The transformation of the G-code into the UAV Trajectory occurs in two distinct steps. Firstly, by assuming an average speed for the UAV all the point segments coming from the G-code are considered to be waypoints for the new trajectory. An interpolation between them takes place by calculating the duration for the UAV to pass through each one and assigning it to this trajectory segment. As a result, the trajectory of the extruder of the UAV is extracted in a form that could be easily interfaced with any controller. Though, it refers to the end-effector of the extruder and not the UAV itself. A transformation needs to be calculated that transforms the trajectory in the body frame of the UAV. In this context, an arm is assumed to be hanging below the UAV with one joint before the end-effector, as shown in Fig. \ref{fig:uav_extruder}.
The transformation matrix $\mathbf{M}^G_B$ for the above can be found in Eq.\ref{eq:extruder_tf}
\newline
\newline
\newline
\begin{minipage}{.45\columnwidth}
    \begin{equation*}
        \mathbf{M}^G_B=\begin{bmatrix}
            0 & 0 & 0 & 0 \\
            0 & 0 & 0 & 0 \\
            0 & 0 & 0 & -l_{ex} \\ne
            0 & 0 & 0 & 1 \\
        \end{bmatrix}
    \end{equation*}
\end{minipage}%
\begin{minipage}{.45\columnwidth}
    \begin{equation*}
        \mathbf{M}^E_G=\begin{bmatrix}
            c\theta & 0 & s\theta & 0 \\
            0 & 1 & 0 & 0 \\
            -s\theta & 0 & c\theta & -l_{g} \\
            0 & 0 & 0 & 1 \\
        \end{bmatrix}
    \end{equation*}
\end{minipage}
\vspace{0.3cm}
\begin{equation}\label{eq:extruder_tf}
    \mathbf{M}^G_B = {\mathbf{M}^E_G}^{-1} * {\mathbf{M}^G_B}^{-1}     
\end{equation}

\subsection{UAV Control}
After having been transformed into the body frame of the robot, the trajectory is fed to the controller of the UAV. The controller is a nonlinear tracking controller developed
on the special Euclidean group SE(3) \cite{lee_controller}. Its ability to both track a desired trajectory and a desired position is suitable for the application. The disturbances caused by the manipulator hanging from the UAV are not taken into account on the model of the controller but assumed as external disturbances that the controller has to compensate for. This can be considered a valid assumption since there are no abrupt movements and no big velocities are demanded throughout the execution that would lead to the swinging action of the extruder.

\section{SIMULATION RESULTS} \label{Sec:Simulation Results}
\begin{figure}[H]
    \begin{center}
        \includegraphics[width=0.7\columnwidth]{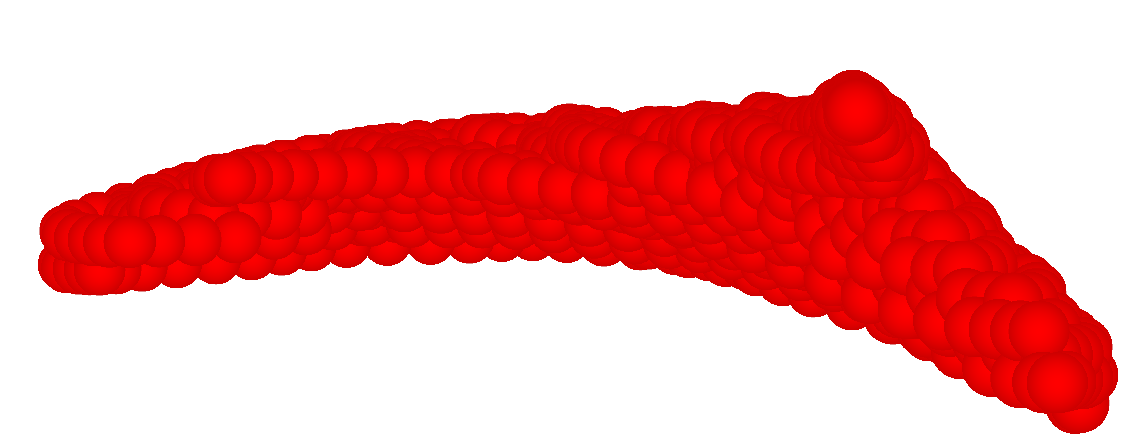}
        \caption{One chunk printed by a UAV. Material deposition is simulated by spheres stacked together.}
        \label{fig:simplePrint}
    \end{center}
\end{figure}

\begin{figure}[H]
    \centering
    \includegraphics[width=0.7\columnwidth]{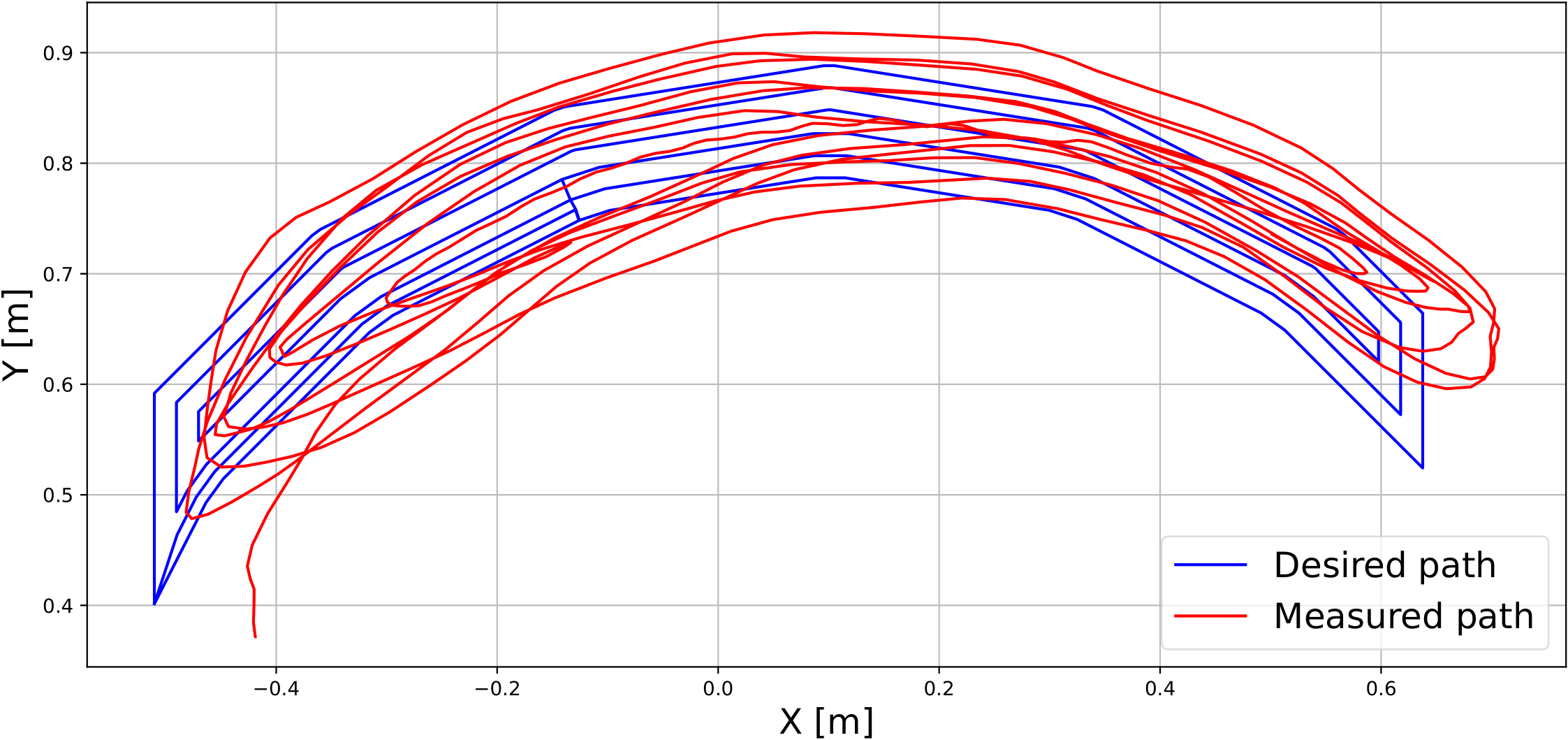}
    \caption{Desired (blue) and Measured (red) UAV Path on one Layer of the print}
    \label{fig:2D_error}
\end{figure}

The experiments for this framework were carried out in the simulation environment of Gazebo and the package "rotors" \cite{rotors_simulator} was used as the simulation framework. Specifically, the model of the "CARMA" \cite{carma} UAV was selected and had no actuation on the robotic arm hanging below, so the angle of $\phi=0 \degree$. In order to simulate the deposition of the material, an odometry sensor is attached to the end-effector of the arm and is kept track of while the UAV is moving. The printing action starts when the command is given by the UAV and the material is visualized by small spheres being deposited below the extruder's nozzle. The physics of the material being deposited is not modeled in this approach and the assumption of the material will stick to the previous layer is made. The initial search is carried out in a centralized entity and a mission planner that orchestrates the whole printing mission assigns UAVs with chunks to print and imposes their states. The mesh given to the planner is the same as the one of Fig. \ref{fig:search_results}, its original volume is $25.24$L and the UAVs available to print it are eight in total with each one carrying $4$L of material. The final chunks are nine and their volumes range from $1,32$L to $3.85$L. The printing of one chunk can be seen in Fig. \ref{fig:simplePrint} where the consecutive layers of material have been deposited on top of one another. 
The big diameter of the deposited material is enough to compensate for some trajectory tracking errors shown in Fig. \ref{fig:2D_error}. The error is observed to be bigger around segments of the part that the UAV needs to go towards a different direction which causes a momentary swinging motion of the extruder and a small rotation of the UAV leads to a big position error of the extruder due to its length.
        
%

%
%

\begin{figure}[]
    \begin{center}
        \includegraphics[width=\textwidth]{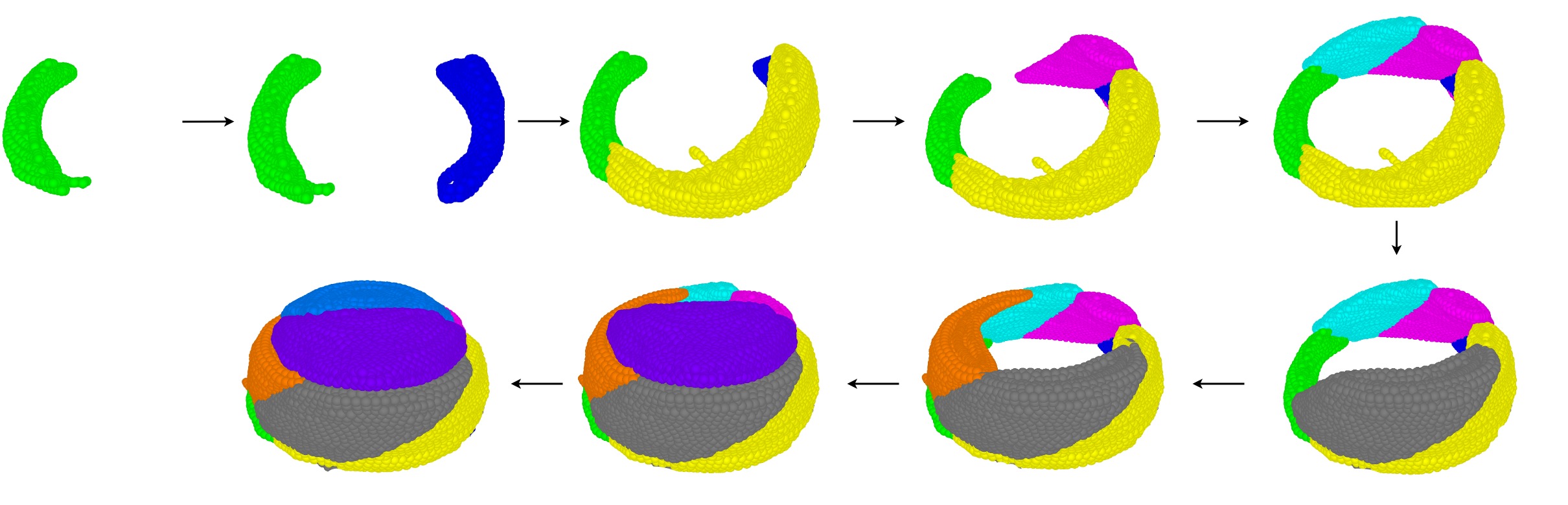}
         \caption{Sequential simulation snaphsots of chunks being printed with different colors} 
        \label{fig:sequential}
    \end{center}
\end{figure}

\begin{figure}[H]
    \begin{center}\label{fig:sim_final}
        \includegraphics[width=0.6\columnwidth]{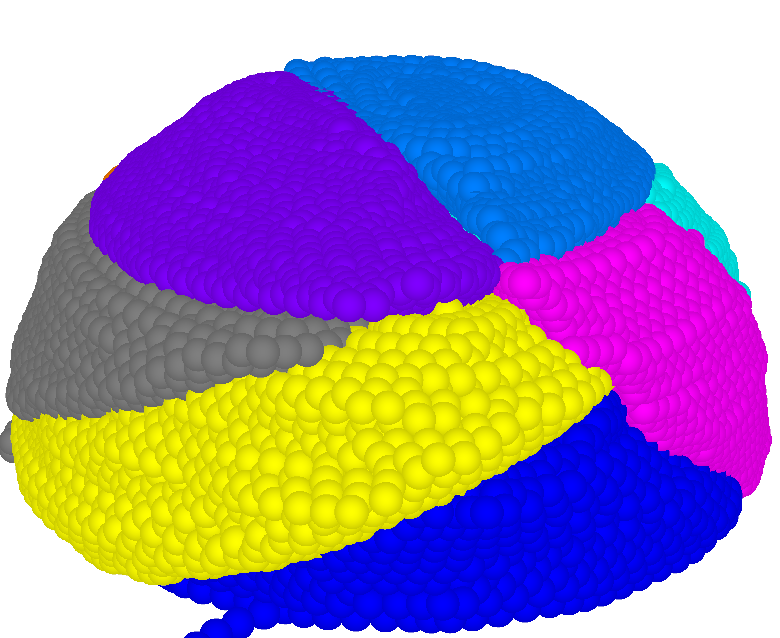}
        \caption{Final mesh composed of all the chunks generated and printed sequentially from the available UAVs}
        \label{fig:multiplePartsSim}
    \end{center}
\end{figure}

Following the same logic as before, the printing of all the chunks takes place. Each chunk is printed sequentially by its assigned UAV based on the priority generated from the topology of the BSP tree. Multiple snapshots of the printing process can be seen in Fig. \ref{fig:sequential}, where each chunk is printed one by one following the priority given by the BSP tree. The final result can be seen in Fig. \ref{fig:multiplePartsSim}

%
\section{Conclusions} \label{sec:conclusions}
In this article, a flexible multi-DoF large-scale aerial 3D printing framework, supported with automated optimal slicing, has been presented. The proposed framework 
introduces a generic planner slicing mechanism that optimally decomposes an arbitrary geometric shape into multiple printable chunks. The chunks are constructed by incorporating the constraint imposed based on the shape of the extruder, thereby eliminating the possibility of collision  extruder, while depositing material over the previously printed shape. In addition, an innovative task scheduling mechanism is presented that sets the basis to enable seamless integration of distributed aerial $3D$ printing with multiple UAVs in multiaxis-layered manufacturing. The proposed aerial 3D printing framework is demonstrated with a simulation performed in Gazebo Simulator.



\clearpage
\bibliographystyle{IEEEtran}
\bibliography{sample}

\end{document}